\documentclass[10pt]{article}
\usepackage[a4paper, total={6in, 8in}]{geometry}

\usepackage{breakcites}
\usepackage[utf8]{inputenc} 
\usepackage[T1]{fontenc}    
\usepackage{hyperref}       
\usepackage{url}            
\usepackage{booktabs}       
\usepackage{amsfonts}       
\usepackage{nicefrac}       
\usepackage{microtype}      

\usepackage{siunitx}
\usepackage{textcomp}
\usepackage{graphicx}
\usepackage{amsmath,amssymb}
\usepackage{enumitem}
\usepackage{subcaption}
\usepackage{bbm}
\usepackage{algorithm2e}
\usepackage{algorithm,algorithmic}
\usepackage{varwidth}
\newcommand{\approach}{\emph{CNC}}

\title{Generalized Clustering by Learning to Optimize\\ Expected Normalized Cuts}

\author
{
Azade Nazi$^{\dag}$,
Will Hang$^{\dag\dag\dag}$,
Anna Goldie$^{\dag}$,
Sujith Ravi$^{\dag\dag}$, 
Azalia Mirhoseini$^{\dag}$
\\
$^{\dag}$Google Brain;
$^{\dag\dag}$Google Research;
$^{\dag\dag\dag}$Stanford University
\\
\small{\{azade,~agoldie,~sravi,~azalia\}@google.com},
\small{\{willhang\}@stanford.edu}
}
\date{}

\begin{document}

\maketitle

\begin{abstract}

We introduce a novel end-to-end approach for learning to cluster in the absence of labeled examples.
Our clustering objective is based on optimizing normalized cuts, a criterion which measures both intra-cluster similarity as well as inter-cluster dissimilarity. We define a differentiable loss function equivalent to the expected normalized cuts. Unlike much of the
work in unsupervised deep learning, our trained model directly outputs final cluster assignments, rather than embeddings that need further processing to be usable. Our approach generalizes to unseen datasets across a wide variety of domains, including text, and image. Specifically, we achieve state-of-the-art results on popular unsupervised clustering benchmarks (e.g., MNIST, Reuters, CIFAR-10, and CIFAR-100), outperforming the strongest baselines by up to 10.9\%. Our generalization results are superior (by up to 21.9\%) to the recent top-performing clustering approach with the ability to generalize. 


\end{abstract}

\section{Introduction}
\label{sec:intro}

Clustering unlabeled data is an important problem from both a scientific and practical perspective. As technology plays a larger role in daily life, the volume of available data has exploded. However, labeling this data remains very costly and often requires domain expertise. Therefore, unsupervised clustering methods are one of the few viable approaches to gain insight into the structure of these massive unlabeled datasets. 

One of the most popular clustering methods is spectral clustering \cite{Shi_2000, ng2002spectral, von2007tutorial}, which first embeds the similarity of each pair of data points in the Laplacian's eigenspace and then uses k-means to generate clusters from it. Spectral clustering not only outperforms commonly used clustering methods, such as k-means \cite{von2007tutorial}, but also allows us to directly minimize the pairwise distance between data points and solve for the optimal node embeddings analytically. Moreover, it is shown that the eigenvector of the normalized Laplacian matrix can be used to find the approximate solution to the well known normalized cuts problem~\cite{ng2002spectral, von2007tutorial}.

In this work, we introduce \approach, a framework for \textit{\textbf{C}lustering} by learning to optimize expected \textit{\textbf{N}ormalized \textbf{C}uts}. We show that by directly minimizing a continuous relaxation of the normalized cuts problem, \approach\ enables end-to-end learning approach that outperforms top-performing clustering approaches. We demonstrate that our approach indeed can produce lower normalized cut values than the baseline methods such as SpectralNet, which consequently results in better clustering accuracy.

\begin{figure*}[t]
\centering
 \includegraphics[width=140mm,height=45mm]{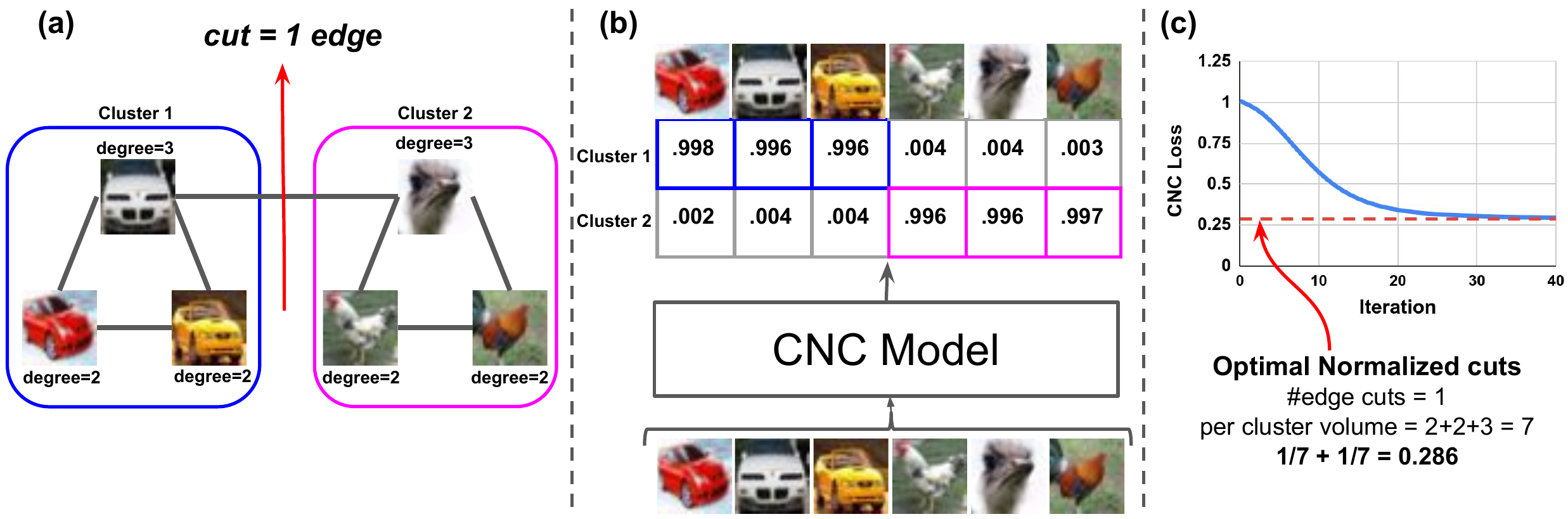}
 \vspace{-0.1in}
    \caption{\small{Motivational example: (a) affinity graph of 6 images from CIFAR-10, the objective is to cluster these images into two clusters. (b) \approach\ model is trained to minimize expected normalized cuts in an unsupervised manner without the need for any labeled data. For each data point, our model directly outputs the probabilities of it belonging to each of the clusters. (c) The \approach\ loss converges to the optimal normalized cuts value. In Algorithm \ref{alg:CNC} we show how we can scale this approach through a batch processing technique to large datasets.}}
    \label{fig:motiv}
\end{figure*}

Let us motivate \approach\ through a simple example. In Figure~\ref{fig:motiv}, we want to cluster 6 images from CIFAR-10 dataset into two clusters. The affinity graph for these data points is shown in Figure~\ref{fig:motiv}(a) (details of constructing such graph is discussed in Section~\ref{sec:embedding}). In this example, it is obvious that the optimal clustering is the result of cutting the edge connecting the two triangles. Cutting this edge will result in the optimal value for the normalized cuts objective. In \approach, we define a new differentiable loss function equivalent to the expected normalized cuts objective. We train a deep learning model to minimize the proposed loss in an unsupervised manner without the need for any labeled datasets. Our trained model directly returns the probabilities of belonging to each cluster (Figure~\ref{fig:motiv}(b)). In this example, the optimal normalized cuts is 0.286 (Equation~\ref{equ:Ncut}), and as we can see, the \approach\ loss also converges to this value (Figure~\ref{fig:motiv}(c)).



We compare the performance of \approach\ to several learning-based clustering approaches (SpectralNet~\cite{shaham2018spectralnet}, DEC~\cite{xie2016unsupervised}, DCN~\cite{yang2017}, VaDE~\cite{zheng2016variational}, DEPICT~\cite{Dizaji2017}, IMSAT~\cite{Weihua2017}, and IIC~\cite{xuIIC2018}) on four datasets: MNIST, Reuters, CIFAR10, and CIFAR100. Our results show up to 10.9\% improvement over the baselines. Moreover, generalizing spectral embeddings to unseen data points, a task commonly referred to as out-of-sample-extension (OOSE), is a non-trivial task~\cite{Bengio:2003, Belkin:2006, MendozaQuispe:2016}. Our results confirm that \approach\ generalizes to unseen data. Our generalization results are superior (by up to 21.9\%) to SpectralNet~\cite{shaham2018spectralnet}, the recent top-performing clustering approach with the ability to generalize. 


\section{Related Work}
\label{sec:relWork}
Recent deep learning approaches to clustering attempt to embed the input data into a form that is amenable to clustering by k-means or Gaussian Mixture Models.
\cite{yang2017, xie2016unsupervised} focused on learning representations for clustering. 
To find the clustering-friendly latent representations and to better cluster the data, DCN~\cite{yang2017} proposed a joint dimensionality reduction (DR) and K-means clustering approach in which DR is accomplished via learning a deep neural network.
DEC \cite{xie2016unsupervised} simultaneously learns cluster assignment and the underlying feature representation by iteratively updating a target distribution to sharpen cluster associations. 

Several other approaches rely on a variational autoencoder that utilizes a Gaussian mixture prior~\cite{zheng2016variational, dilokthanakul2016deep, Weihua2017, xuIIC2018, Matan2018}. These approaches are mainly based on data augmentation, where the network is trained to maximize the mutual information between inputs and predicted clusters, while regularizing the network so that the cluster assignment of the data points is consistent with the assignment of the augmented points.

Different clustering objectives, such as self-balanced k-means and balanced min-cut, have also been exhaustively studied~\cite{liu_2017, chen_2017, chang_2014}. One of the most effective techniques is spectral clustering, which first generates node embeddings in the eigenspace of the graph Laplacian, and then applies k-means clustering to these vectors \cite{Shi_2000, ng2002spectral, von2007tutorial}. To address the fact that clusters with the lowest graph conductance tend to have few nodes~\cite{LeskovecLDM09, ZhangR18}, ~\cite{ZhangR18} proposed regularized spectral clustering to encourage more balanced clusters.

Generalizing clustering to unseen nodes and graphs is nontrivial~\cite{Bengio:2003, Belkin:2006, MendozaQuispe:2016}. A recent work, SpectralNet \cite{shaham2018spectralnet}, takes a deep learning approach to spectral clustering that generalizes to unseen data points. This approach first learns embeddings of the similarity of each pair of data points in Laplacian’s eigenspace and then applies k-means to those embeddings to generate clusters. Unlike SpectralNet, we propose an end-to-end learning approach with a differentiable loss that directly minimizes the normalized cuts. We show that our approach indeed can produce lower normalized cut values than the baseline methods such as SpectralNet, which consequently results in better clustering accuracy. Our evaluation results show that \approach\ improves generalization accuracy on unseen data points by up to 21.9\%. 

\section{Preliminaries}
Since \approach\ objective is based on optimizing normalized cuts, in this section, we briefly overview the formal definition of this metric.

\subsection{Formal definition of Normalized cuts}
\label{sec:formalncut}

Let $G=(V,E, W)$ be a graph where $V = \{v_i\}$ and $E = \{e(v_i, v_j) | v_i \in V, v_j \in V\}$ are the set of nodes and edges in the graph and $w_{ij} \in W$ is the edge weight of the $e(v_i, v_j)$. Let $n$ be the number of nodes. A graph $G$ can be clustered into $g$ disjoint sets $S_1, S_2, \dots S_g$, where the union of the nodes in those sets are $V$ ($\bigcup_{k=1}^{g} S_k = V$), and each node belongs to only one set ($\bigcap_{k=1}^{g} S_k = \emptyset$), by simply removing edges connecting those sets. For example, in Figure~\ref{fig:motiv}(a), by removing one edge two disjoint clusters are formed. 

Normalized cuts ($\emph{Ncuts}$) which is defined based on the graph conductance, has been studied by~\cite{Shi_2000, ZhangR18}, and the cost of a cut that forms disjoint sets $S_1, S_2, \dots S_g$ is computed as:

\begin{equation} \label{equ:Ncut}
 \emph{Ncuts}(S_1, S_2, \dots S_g) = \sum_{k=1}^{g} \frac{\emph{cut}(S_k, \bar{S}_k)}{\emph{vol}(S_k, V)}
\end{equation}

Where $\bar{S}_k$ represents the complement of $S_k$, i.e., $\bar{S}_k = \bigcup_{i \neq k} S_i$. $\emph{cut}(S_k, \bar{S}_k)$ is called \emph{cut} and is the total weight of the edges that are removed from $G$ in order to form disjoint sets $S_k$ and $\bar{S}_k$. $\emph{vol}(S_k, V)$ is the total edge weights ($w_{ij}$), whose end points ($v_i$, or $v_j$) belong to $S_k$. The \emph{cut} and \emph{vol} are:

\begin{equation} \label{equ:cut}
\emph{cut}(S_k, \bar{S}_k) = \sum_{v_i\in S_k, v_j\in \bar{S}_k} w_{ij} ~~~~~,~~~~
\emph{vol}(S_k, V) = \sum_{v_i\in S_k} \sum_{v_j\in V} w_{ij}
\end{equation}


Note that in Equation~\ref{equ:cut}, $S_k$ and $\bar{S}_k$ are disjoint, i.e., $S_k \cap \bar{S}_k = \emptyset$, while in \emph{vol}, $S_k \subset V$.
In running example (Figure~\ref{fig:motiv}), since the edge weights are one, $\emph{cut}(S_1, \bar{S}_1) = \emph{cut}(S_2, \bar{S}_2) = 1$, and $\emph{vol}(S_1, V) = \emph{vol}(S_2, V) = 2+2+3 = 7 $. Thus the $\emph{Ncuts}(S_1, S_2) = \frac{1}{7} + \frac{1}{7} = 0.286$. In this example one can see that such clustering results in minimum value of the normalized cuts. \approach\ aims to find a cut that the normalized cuts (Equation~\ref{equ:Ncut}) is minimized.
\section{CNC Framework}
\label{lbl:framework}

Finding the cluster assignments that minimizes the normalized cuts is NP-complete and an approximation to the this problem is based on the eigenvectors of the normalized graph Laplacian which has been studied in~\cite{Shi_2000, ZhangR18}. \approach, on the other hand, is a neural network framework for learning to cluster in the absence of labeled examples by directly minimizing the continuous relaxation of the normalized cuts. As shown in Algorithm~\ref{alg:CNC}, end-to-end training of the \approach\ contains two steps, i.e, (i) data points embedding (line 3), and (ii) clustering (lines 4-9). In data points embedding, the goal is to learn embeddings that capture the affinity of the data points, while the clustering step uses those embeddings to learn the CNC model and outputs the cluster assignments. 
Next, we first focus on the clustering step and we introduce our new diffrentiable loss function to train CNC model. Later in Section~\ref{sec:embedding}, we discuss the details of the embedding step.

\begin{algorithm*}[t]
\setlength{\AlCapSkip}{0em}
\caption{\\End-to-End Training of \approach: \textit{\textbf{C}lustering} by learning to optimize expected \textit{\textbf{N}ormalized \textbf{C}uts}}
\begin{algorithmic} [1]
\label{alg:CNC}
\STATE{} {{\bf Input: } dataset $X \in \mathbb{R}^{m}$, number of clusters $g$, data point embedding size $d$, batch size $m$}
\STATE {{\bf Output: } Cluster assignments of data points.}
\\ {\bf \textit {Preprocessing step, learn data points embedding (details in Section~\ref{sec:embedding})}:}\\
\STATE {Given a dataset $X = \{x_1,\dots x_n\}$, train a Siamese network to find embeddings $\{v_1,\dots v_n\},~~v_i \in \mathbb{R}^{d}$ that represent the affinity of the data points. $G_{\theta_{\text{siamese}}}: \mathbb{R}^{m} \rightarrow \mathbb{R}^{d}$}
\\ {\bf \textit {Clustering step, learn CNC model $F_{\theta}$ (details in Section~\ref{sec:loss})}:}\\
\WHILE {CNC loss in Equation~\ref{equ:YNcut} not converged}
    \STATE{Sample a random minibatch $M$ of size $m$ from the embeddings}
    \STATE{Compute affinity graph $W \in  \mathbb{R}^{m \times m}$ over the $M$ based on the k-nearest neighbors}
    \STATE{Use $M$ and $W$ to train CNC model $F_{\theta}: \mathbb{R}^{d} \rightarrow \mathbb{R}^{g}$ that minimizes the expected normalized cuts (Equation ~\ref{equ:YNcut}). For a data point with embedding $v_i$ the output $y_i = F_{\theta}(v_i)$ represents the assignment probabilities over $g$ clusters.}
\ENDWHILE
\\ {\bf \textit {Inference, cluster assignments}}\\
\STATE {For every data points $x_i$ whose embedding is $v_i$ return $\arg\max$ of $y_i = F_{\theta}(v_i)$ as its cluster assignment.}
\end{algorithmic}
\end{algorithm*}

\subsection{Clustering Step: Learn CNC model}
\label{sec:loss}
In this section, we describe the clustering step in Algorithm~\ref{alg:CNC} (lines 4-9). for each data point $x_i$, the input to clustering step is embedding $v_i \in \mathbb{R}^{d}$ (detail in Section~\ref{sec:embedding}). The goal is to learn CNC model $F_{\theta}: \mathbb{R}^{d} \rightarrow \mathbb{R}^{g}$ that for a given embedding $v_i \in \mathbb{R}^{d}$ it returns $y_i = F_{\theta}(v_i) \in \mathbb{R}^{g}$, which represents the assignment probabilities over $g$ clusters. Clearly for $n$ data points, it returns $Y \in \mathbb{R}^{n \times g}$ where $Y_{ik}$ represents the probability that $v_i$ belongs to cluster $S_k$. The CNC model $F_{\theta}$ is implemented using a neural network, where the parameter vector $\theta$ denotes the network weights. We propose a loss function based on output $Y$ to calculate the expected normalized cuts. Thus \approach\ learns the $F_{\theta}$ by minimizing this loss.

Recall that $\emph{cut}(S_k, \bar{S}_k)$ is the total weight of the edges that are removed from $G$ in order to form disjoint sets $S_k$ and $\bar{S_k}$. In our setup, embeddings are the nodes in graph $G$, and neighbors of an embedding $v_i$ are based on the k-nearest neighbors. Let $Y_{ik}$ be the probability that node $v_i$ belongs to cluster $S_k$. The probability that node $v_j$ does not belong to $S_k$ would be $1 - Y_{jk}$. Therefore, $\mathbb{E}[\emph{cut}(S_k, \bar{S_k})]$ can be formulated by Equation~\ref{equ:Ycut1}, where $\mathcal{N}(v_i)$ is the set of nodes adjacent to $v_i$.
%
\begin{equation}
\label{equ:Ycut1}
\begin{aligned}
\mathbb{E}[\emph{cut}(S_k, \bar{S}_k)] &= \sum_{\substack{v_i \in S_k \\ v_j \in \mathcal{N}(v_i)}}  w_{ij} \sum_{z=1}^{g} Y_{iz} (1-Y_{jz})
\end{aligned}
\end{equation}
Since the weight matrix $W$ represents the edge weights adjacent nodes, we can rewrite Equation~\ref{equ:Ycut1}:
%
\begin{equation}
\label{equ:Ycut}
\begin{aligned}
\mathbb{E}[\emph{cut}(S_k, \bar{S}_k)] &= \sum_\text{reduce-sum} Y_{:,k} (1-Y_{:,k})^\intercal \odot W
\end{aligned}
\end{equation}

The element-wise product with the weight matrix $(\odot \ W)$ ensures that only the adjacent nodes are considered. Moreover, the result of $Y_{:,k} (1-Y_{:,k})^\intercal \odot W$ is an $n \times n$ matrix and \emph{reduce-sum} is the sum over all of its elements.

From Equation~\ref{equ:cut}, $\emph{vol}(S_k, V)$ is the total edge weights ($w_{ij}$), whose end points ($v_i$, or $v_j$) belong to $S_k$. Let $D$ be a column vector of size $n$ where $D_i$ is the total edge weights from node $v_i$. Given $Y$, we can calculate the $\mathbb{E}[\emph{vol}(S_k, V)]$ as follows:
%
\begin{equation}
\label{equ:assoc}
\begin{aligned}
\Gamma = Y^\intercal D \\
\mathbb{E}[\emph{vol}(S_k, V)] &= \Gamma_k
\end{aligned}
\end{equation}

where $\Gamma$ is a vector in $\mathbb{R}^g$, and $g$ is the number of sets/clusters. 

With $\mathbb{E}[\emph{cut}(S_k, \bar{S}_k)]$ and $\mathbb{E}[\emph{vol}(S_k, V)]$ from Equations~\ref{equ:Ycut} and~\ref{equ:assoc}, we can calculate the expected normalized cuts as follows:
%
\begin{equation}
\label{equ:YNcut}
\begin{aligned}
\mathbb{E}[\emph{Ncuts}(S_1, S_2, \dots S_g)] &= \sum_\text{reduce-sum} (Y \oslash \Gamma) (1-Y)^\intercal \odot W
\end{aligned}
\end{equation}

$\oslash$ is element-wise division and the result of $(Y \oslash \Gamma) (1-Y)^\intercal \odot W$ is a $n \times n$ matrix where \emph{reduce-sum} is the sum over all of its elements.

As you can see the affinity graph $W$ is part of the \approach\ loss (Equation~\ref{equ:YNcut}). Clearly, when the number of data points ($n$) is large, such calculation can be expensive. However, in our experimental results, we show that for large dataset (e.g., Reuters contains 685,071 documents), it is possible to optimize the loss on randomly sampled minibatches of data. We also build the affinity graph over a given minibach using the embeddings and based on their k nreast-neighbor (Algorithm~\ref{alg:CNC} (lines 5-6)). Specifically, in our implementation, \approach\ model $F_{\theta}$ is a fully connected layer followed by gumble softmax, trained on randomly sampled minibatches of data to minimize Equation~\ref{equ:YNcut}. When training is over, the final assignment of a node $v_i$ to a cluster is the $\arg\max$ of $y_i = F_{\theta}(v_i)$ (Algorithm~\ref{alg:CNC} (line 9)).

\subsection{Embedding Step}
\label{sec:embedding}
In this section, we discuss the embedding step (line 3 in Algorithm~\ref{alg:CNC}). Different affinity measures, such as simple euclidean distance or nearest neighbor pairs combined with a Gaussian kernel, have been used in spectral clustering. Recently it is shown that unsupervised application of a Siamese network to determine the distances improves the quality of the clustering~\cite{shaham2018spectralnet}. 

In this work, we also use Siamese networks to learn embeddings that capture the affinities of the data points.
Siamese network is trained to learn an adaptive nearest neighbor metric. It learns the affinities directly from euclidean proximity by ''labeling'' points $x_i$, $x_j$ positive if $\|x_i - x_j\|$ is small and negative otherwise.
In other words, it generates embeddings such that adjacent nodes are closer in the embedding space and non-adjacent nodes are further. Such network is typically trained to minimize contrastive loss:
$$
L_{\text{siamese}} = 
\begin{cases}
  ||v_i - v_j||^2,  & (x_i, x_j)\text{ is a positive pair}\\    
  \max(1 - ||v_i - v_j||^2, 0)^2, & (x_i, x_j)\text{ is a negative pair} 
\end{cases}
$$

where $v_i = G_{\theta_{\text{siamese}}}(x_i)$, and $G_{\theta_{\text{siamese}}}: \mathbb{R}^{m} \rightarrow \mathbb{R}^{d}$ is a Siamese network that transforms representations in the input space $x_i \in \mathbb{R}^{m}$ to embeddings $v_i \in \mathbb{R}^{d}$.

\section{Experiments}
\label{sec:exp}
The main goals of our experiments are to evaluate: (a) 
The performance of \approach\ against the existing clustering approaches. (b) The ability of \approach\ to generalize to unseen data compared to the top-performing generalizable baseline. (c) The effectiveness of minimizing Normalized cuts on improving the clustering results. (d) The generalization performance of \approach\ as we vary the number of data points in training dataset.

\subsection{Datasets and Baseline Methods}
We evaluate the performance of \approach\ in comparison to several deep learning-based clustering approaches on four real world datasets: MNIST, Reuters, CIFAR-10, and CIFAR-100. The details of the datasets are as follows:
\begin{itemize}[leftmargin=*]
\item{MNIST is a collection of 70,000 28×28 gray-scale images of handwritten digits, divided into 60,000 training images and 10,000 test images.}
\item{The Reuters dataset is a collection of English news labeled by category. Like SpectralNet, DEC, and VaDE, we used the following categories: corporate/industrial, government/social, markets, and economics
as labels and discarded all documents with multiple labels. Each article is represented by a tfidf vector using the 2000 most frequent words. The dataset contains 685,071 documents. We divided the data randomly to a 90\%-10\%
split to evaluate the generalization ability of \approach.
We also investigate the imapact of training data size on the generalization by considering following splits: 90\%-10\%, 70\%-30\%, 50\%-50\%, 20\%-80\%, and 10\%-90\%.}
\item{CIFAR-10 consists of 60000 32x32 colour images in 10 classes, with 6000 images per class. There are 50000 training images and 10000 test images.}
\item{CIFAR-100 has 100 classes containing 600 images each with a 500/100 train/test split per class.}
\end{itemize}

In all runs we assume the number of clusters is given. In MNIST and CIFAR-10 number of clusters (g) is 10, g = 4 in Reuters, g = 100 in CIFAR-100. We compare \approach\ to SpectralNet~\cite{shaham2018spectralnet}, DEC~\cite{xie2016unsupervised}, DCN~\cite{yang2017}, VaDE~\cite{zheng2016variational}, DEPICT~\cite{Dizaji2017}, IMSAT~\cite{Weihua2017}, and IIC~\cite{xuIIC2018}. While \cite{yang2017, xie2016unsupervised} focused on learning representations for clustering, other approaches~\cite{zheng2016variational, dilokthanakul2016deep, Weihua2017, xuIIC2018, Matan2018} rely on a variational autoencoder that utilizes a Gaussian mixture prior. SpectralNet \cite{shaham2018spectralnet}, takes a deep learning approach to spectral clustering that generalizes to unseen data points. Table \ref{tbl:clustering} shows the results reported for these six methods.

Similar to~\cite{shaham2018spectralnet}, for MNIST and Reuters we use publicly available and pre-trained autoencoders$\footnote{\url{https://github.com/slim1017/VaDE/tree/master/pretrain_weights}}$. The autoencoder used to map the Reuters data to code space was trained based on a random subset of 10,000 samples from the full dataset. Similar to~\cite{Weihua2017}, for CIFAR-10 and CIFAR-100 we applied 50-layer pre-trained deep residual networks trained on ImageNet to extract features and used them
for clustering.

\subsection{Performance Measures}
We use two commonly used measures, the unsupervised clustering accuracy (ACC), and the normalized mutual information (NMI) in~\cite{Cai2011} to evaluate the accuracy of the clustering. Both ACC and NMI are in [0, 1], with higher values indicating better correspondence the clusters and the true labels. Note that true labels never used neither in training, nor in test. 

\noindent\emph{Clustering Accuracy (ACC):}
For data points $X = \{x_1, \dots x_n\}$, let $l = (l_1, \dots l_n)$ and $c = (c_1, \dots c_n)$ be the true labels and predicted clusters respectively. The ACC is defined as:
\begin{equation*}
\label{equ:Acc}
\begin{aligned}
ACC(l,c) = \frac{1}{n} \max_{\pi \in \prod} \sum_{i=1}^{n}  \mathbbm{1} \{l_i = \pi(c_i)\}
\end{aligned}
\end{equation*}
where {\small$\prod$} is the collection of all permutations of ${1, \dots g}$. The optimal permutation $\pi $ can be computed using the Kuhn-Munkres algorithm~\cite{Munkres1957Assignment}. 

\noindent\emph{Normalized Mutual Information (NMI):}
Let $I(l; c)$ be the mutual information between $l$ and $c$, and $H$(.) be their entropy. The NMI is:
\begin{equation*}
\label{equ:NMI}
\begin{aligned}
NMI(l,c) = \frac{I(l;c)}{\max \{H(l),H(c)\}} 
\end{aligned}
\end{equation*}

\subsection{Experimental Results}
For each dataset we trained a Siamese network~\cite{Hadsell:2006, Shaham2018LearningBC} to learn embeddings which represents the affinity of data points by only considering the k-nearest neighbors of each data. In Table~\ref{tbl:clustering}, we compare clustering performance across four benchmark datasets. Since most of the clustering approaches do not generalize to unseen data points, all data has been used for the training (Later in Section~\ref{sec:gen}, to evaluate the generalizability we use 90\%-10\% split for training and testing).

\begin{table*}
  \centering
  \begin{tabular}{lllllllll}
    \toprule
    & \multicolumn{2}{c}{MNIST} & \multicolumn{2}{c}{Reuters} & \multicolumn{2}{c}{CIFAR-10} & \multicolumn{2}{c}{CIFAR-100} \\
    \cmidrule(r){2-3}
    \cmidrule(r){4-5}
    \cmidrule(r){6-7}
    \cmidrule(r){8-9}
    Method & ACC & NMI & ACC & NMI & ACC & NMI & ACC & NMI \\
    \midrule
    DEC & 0.843$^{*}$ & 0.800$^{*}$ & 0.756$^{*}$ & - & 0.469 & - & - & - \\
    DCN & 0.830$^{**}$ & 0.810$^{**}$ & - & - & - & - & - & - \\
    VaDE & 0.945$^{\dag}$ & - & 0.794$^{\dag}$ & - & - & - & - & - \\
    DEPICT & 0.965$^{\dag\dag}$ & 0.917$^{\dag\dag}$ & - & - & - & - & - & - \\
    IMSAT & 0.984$^{\ddag\ddag}$ & - & 0.719$^{\ddag\ddag}$ & - & 0.456$^{\ddag\ddag}$ & - & 0.275$^{\ddag\ddag}$ & - \\
    IIC & \textbf{0.993}$^{\dag\dag\dag}$ & - & - & - & 0.617$^{\dag\dag\dag}$ & - & - & - \\
    SpectralNet & 0.971$^{\ddag}$ & \textbf{0.924}$^{\ddag}$ & 0.803$^{\ddag}$ & 0.532$^{\ddag}$ & 0.501 & 0.463 & 0.236 & 0.231 \\
    \textbf{\approach} & 0.972 & \textbf{0.924} & \textbf{0.824} & \textbf{0.583} & \textbf{0.702} & \textbf{0.586} & \textbf{0.345} & \textbf{0.502} \\
    \bottomrule
  \end{tabular}
  \caption{Performance of various clustering methods on MNIST, Reuters, CIFAR-10 and CIFAR-100. The model is trained on all data. (--) means values are not reported. ($*$) reported in DEC~[\cite{xie2016unsupervised}], ($**$) reported in DCN~[\cite{yang2017}], ($\dag$) reported in VaDE~[\cite{zheng2016variational}], ($\dag\dag$) reported in DEPICT~[\cite{Dizaji2017}], ($\ddag\ddag$) reported in IMSAT~[\cite{Weihua2017}], ($\ddag$) reported in SpectralNet~[\cite{shaham2018spectralnet}], ($\dag\dag\dag$) reported in IIC~[\cite{xuIIC2018}]. }
\label{tbl:clustering}
\end{table*}

While the improvement of \approach\ is marginal over MNIST, it performs better across other three datasets.
Specifically, over CIFAR-10, \approach\ outperforms SpectralNet and IIC on ACC by 20.1\% and 10.9\% respectively. Moreover, the NMI is improved by 12.3\%.
The results over Reuters, and CIFAR-100, show 0.021\% and 11\% improvement on ACC. The NMI is also 27\% better over CIFAR-100. The fact that our \approach\ outperforms existing approaches in most datasets suggests the effectiveness of using our deep learning approach to optimize normalized cuts for clustering. 

\begin{table*}
  \centering
  \begin{tabular}{lllllllll}
    \toprule
    & \multicolumn{2}{c}{MNIST} & \multicolumn{2}{c}{Reuters} & \multicolumn{2}{c}{CIFAR-10} & \multicolumn{2}{c}{CIFAR-100} \\
    \cmidrule(r){2-3}
    \cmidrule(r){4-5}
    \cmidrule(r){6-7}
    \cmidrule(r){8-9}
    Method & ACC & NMI & ACC & NMI & ACC & NMI & ACC & NMI \\
    \midrule
    SpectralNet & 0.970$^{\ddag}$ & 0.925$^{\ddag}$ & 0.798$^{\ddag}$ & 0.536$^{\ddag}$ & 0.491 & 0.478 & 0.229 & 0.230 \\
    \textbf{\approach} & \textbf{0.971} & \textbf{0.925} & \textbf{0.824} & \textbf{0.586} & \textbf{0.701} & \textbf{0.585} & \textbf{0.343} & \textbf{0.526} \\
    \bottomrule
  \end{tabular}
  \caption{Generalization of clustering methods on MNIST, Reuters, CIFAR-10 and CIFAR-100 datasets. The model is trained only on training set and the reported numbers are the test accuracy. (--) means values are not reported. ($\ddag$) reported in SpectralNet~[\cite{shaham2018spectralnet}].}
  \label{tbl:clustering_gen}
\end{table*}

\begin{table*}
 \centering
    \begin{tabular}{lllll}
    \toprule
    & \multicolumn{1}{c}{MNIST} & \multicolumn{1}{c}{Reuters} & \multicolumn{1}{c}{CIFAR-10} & \multicolumn{1}{c}{CIFAR-100} \\
    \midrule
    SpectralNet & 0.913 & 0.351 & 4.229 & 82.831 \\
    \textbf{\emph{CNC}} & \textbf{0.879} & \textbf{0.21} & \textbf{2.451} & \textbf{58.535} \\
    \bottomrule
    \end{tabular}
    \caption{\small{Numerical value of the normalized cut (Equation~\ref{equ:Ncut}) over the clustering results of the \approach\ and SpectralNet~[\cite{shaham2018spectralnet}]. \approach\ is able to find better cuts than the SpectralNet}}
    \label{tbl:normalized cut}
\end{table*}

\begin{figure*}[h]
\begin{minipage}{0.48\linewidth}
\centering
 \includegraphics[width=65mm,height=35mm]{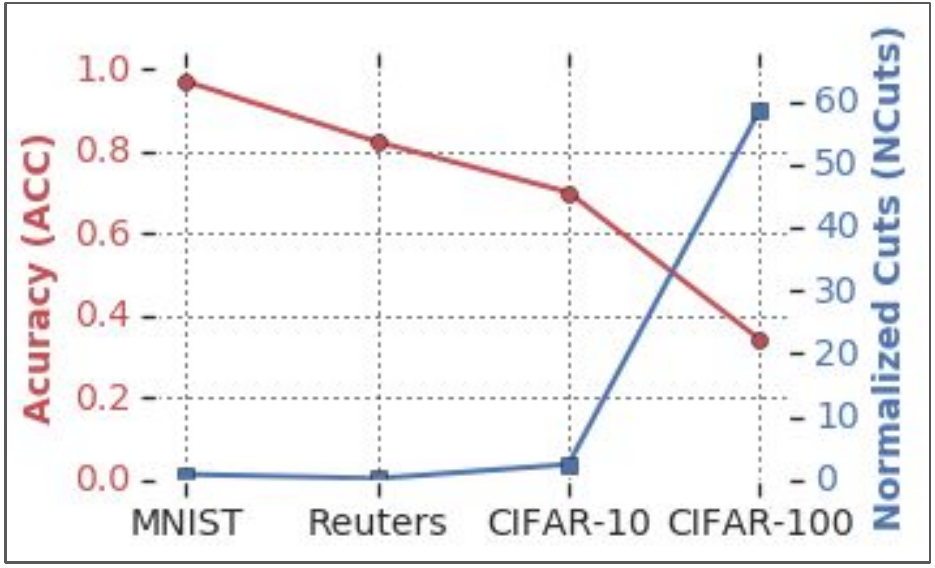}
    \caption{\small{Normalized cuts (right axis) and clustering accuracy (left axis) are anti-correlated. Lower normalized cuts results in better accuracy.}}
    \label{fig:correlation}
\end{minipage}
\hfill
\begin{minipage}{0.48\linewidth}
\centering
 \includegraphics[width=65mm,height=35mm]{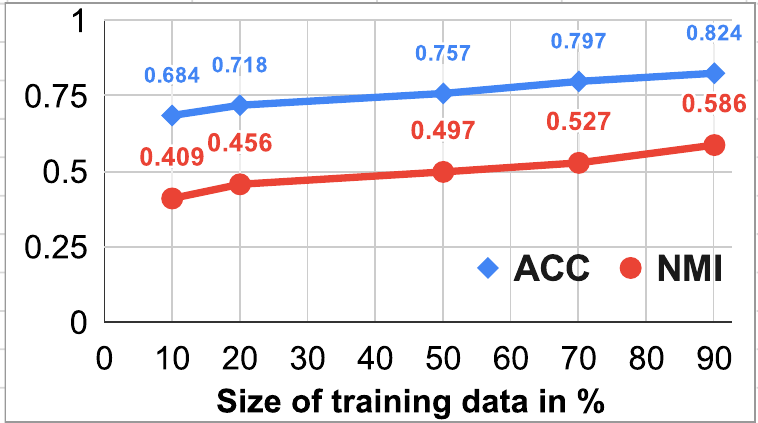}
    \caption{\small{Reuters: with only 10\% training data the ACC and NMI of \approach\ are only 14\% and 18\% lower than ACC and NMI with 90\% training data.}}
    \label{fig:data_size}
\end{minipage}
\end{figure*}

\subsection{Generalization}
\label{sec:gen}
We further evaluate the  generalization ability of \approach\ by dividing the data randomly to a 90\%-10\% split and training on the training set and report the ACC and NMI on the test set (Table~\ref{tbl:clustering_gen}). Among seven methods in Table~\ref{tbl:clustering}, only SpectralNet is able to generalize to unseen data points. \approach\ outperforms SpectralNet in most datasets by up to 21.9\% on ACC and up to 10.7\% on NMI.
Note that simple $\arg\max$ over the output of \approach\ retrieves the clustering assignments while SpectralNet relies on k-means to predict the final clusters.


\subsection{Impact of Normalized cuts in clustering}
To evaluate the impact of normalized cuts for the clustering task, we calculate the numerical value of the Normalized cuts (Equation~\ref{equ:Ncut}) over the clustering results of the \approach\ and SpectralNet. Since such calculation over whole dataset is very expensive we only show this result over the test set.   


Table~\ref{tbl:normalized cut} shows the numerical value of the Normalized cuts over the clustering results of the \approach\ and SpectralNet.
As one can see \approach\ is able to find better cuts than the SpectralNet. Moreover, we observe that for those datasets that the improvement of the \approach\ is marginal (MNIST and Reuters), the normalized cuts of \approach\ are also only slightly better than the SpectralNet, while for the CIFAR-10 and CIFAR-100 that the accuracy improved significantly the normalized cuts of \approach\ are also much smaller than SpectralNet. 
The higher accuracy (ACC in Table~\ref{tbl:clustering_gen}) and smaller normalized cuts (Table~\ref{tbl:normalized cut}), verify that indeed \approach\ loss function is a good notion for clustering task.


\subsection{Imapact of training data size on the generalization}
As you may see in generalization result (Table~\ref{tbl:clustering_gen}), when we reduce the size of the training data to 90\% the accuracy of \approach\ slightly changed in compare to training over the whole data (Table~\ref{tbl:clustering}). Based on this observation, we next investigate how varying the size of the training dataset affects the generalization. In other words, how ACC and NMI of test data change when we vary the size of the training dataset. 

We ran experiment over Routers dataset by dividing the data randomly based on the following data splits: 90\%-10\%, 70\%-30\%, 50\%-50\%, 20\%-80\%, and 10\%-90\%. For example, in 10\%-90\%, we train \approach\ over 10\% of the data and we report the ACC and NMI of \approach\ over the 90\% test set.
Figure~\ref{fig:data_size} shows how the ACC and NMI of \approach\ over the test data change as the size of the training data is varied. For example, when the size of the training data is 90\%, the ACC of \approach\ over the test data is 0.824. 

As we expected and shown in Figure~\ref{fig:data_size} the ACC and NMI of \approach\ increased as the size of the training data is increased.
Interestingly, we observed that with only 10\% training data the ACC of \approach\ is 0.68 which is only 14\% lower than the ACC with 90\% training data. Similarly the NMI of \approach\ with 10\% training data is only 18\% lower than the NMI with 90\% training data.
 

 
\subsection{Model Architecture and Hyper-parameters:}
Here are the details of the \approach\ model for each dataset. 
\begin{itemize}[leftmargin=*]
\item{MNIST: The Siamese network has 4 layers sized [1024, 1024, 512, 10] with ReLU. The clustering module has 2 layers sized [512, 512] with a final gumbel softmax layer. Batch sized is 256 and we only consider 3 nearest neighbors to find the embeddings and constructing the affinity graph for each batch. We use Adam with lr = 0.005 with decay 0.5. Temperature starts at 1.5 and the minimum is set to 0.5.}
\item{Reuters: The Siamese network has 3 layers sized [512, 256, 128] with ReLU. The clustering module has 3 layers sized [512, 512, 512] with tanh activation and a final gumbel softmax layer. Batch sized is 128 and we only consider 3 nearest neighbors to find the embeddings and constructing the affinity graph for each batch. We use Adam with lr = 1e-4 with decay 0.5. Temperature starts at 1.5 and the minimum is set to 1.0.} 
\item{CIFAR-10: The Siamese network has 2 layers sized [512, 256] with ReLU. The clustering module has 2 layers sized [512, 512] with tanh activation and a final gumbel softmax layer. Batch sized is 256 and we only consider 2 nearest neighbors to find the embeddings and constructing the affinity graph for each batch. We use Adam with lr = 1e-4 with decay 0.1. Temperature starts at 2.5 and the minimum is set to 0.5.} 
\item{CIFAR-100: The Siamese network has 2 layers sized [512, 256] with ReLU. The clustering module has 3 layers sized [512, 512, 512] with tanh activation and a final gumbel softmax layer. Batch sized is 1024 and we only consider 3 nearest neighbors to find the embeddings and constructing the affinity graph for each batch. We use Adam with lr = 1e-3 with decay 0.5. Temperature starts at 1.5 and the minimum is set to 1.0.} 
\end{itemize}

\vspace{-0.1in}
\section{Conclusion}
\label{sec:conc}
\vspace{-0.1in}
We propose \approach\ (\textit{\textbf{C}lustering} by learning to optimize \textit{\textbf{N}ormalized \textbf{C}uts}), a framework for learning to cluster unlabeled examples. We define a differentiable loss function equivalent to the expected normalized cuts and use it to train \approach\ model that directly outputs final cluster assignments.
\approach\ achieves state-of-the-art results on popular unsupervised clustering benchmarks (MNIST, Reuters, CIFAR-10, and CIFAR-100 and outperforms the strongest baselines by up to 10.9\%. \approach\ also enables generation, yielding up to 21.9\% improvement over SpectralNet~\cite{shaham2018spectralnet}, the previous best-performing generalizable clustering approach.

\bibliography{gp}

\begin{thebibliography}{}

\bibitem[Belkin et~al., 2006]{Belkin:2006}
Belkin, M., Niyogi, P., and Sindhwani, V. (2006).
\newblock Manifold regularization: A geometric framework for learning from
  labeled and unlabeled examples.
\newblock {\em Journal of machine learning research}, 7:2399--2434.

\bibitem[Ben{-}Yosef and Weinshall, 2018]{Matan2018}
Ben{-}Yosef, M. and Weinshall, D. (2018).
\newblock Gaussian mixture generative adversarial networks for diverse
  datasets, and the unsupervised clustering of images.
\newblock {\em CoRR}, abs/1808.10356.

\bibitem[Bengio et~al., 2003]{Bengio:2003}
Bengio, Y., Paiement, J.-F., Vincent, P., Delalleau, O., Roux, N.~L., and
  Ouimet, M. (2003).
\newblock Out-of-sample extensions for lle, isomap, mds, eigenmaps, and
  spectral clustering.
\newblock In {\em Proceedings of the 16th International Conference on Neural
  Information Processing Systems}, pages 177--184.

\bibitem[Cai et~al., 2011]{Cai2011}
Cai, D., He, X., and Han, J. (2011).
\newblock Locally consistent concept factorization for document clustering.
\newblock {\em IEEE Trans. on Knowl. and Data Eng.}, 23(6):902--913.

\bibitem[Chang et~al., 2014]{chang_2014}
Chang, X., Nie, F., Ma, Z., and Yang, Y. (2014).
\newblock Balanced k-means and min-cut clustering.
\newblock {\em arXiv preprint arXiv:1411.6235}.

\bibitem[Chen et~al., 2017]{chen_2017}
Chen, X., Huang, J.~Z., Nie, F., Chen, R., and Wu, Q. (2017).
\newblock A self-balanced min-cut algorithm for image clustering.
\newblock In {\em ICCV}, pages 2080--2088.

\bibitem[Dilokthanakul et~al., 2016]{dilokthanakul2016deep}
Dilokthanakul, N., Mediano, P.~A., Garnelo, M., Lee, M.~C., Salimbeni, H.,
  Arulkumaran, K., and Shanahan, M. (2016).
\newblock Deep unsupervised clustering with gaussian mixture variational
  autoencoders.
\newblock {\em arXiv preprint arXiv:1611.02648}.

\bibitem[Ghasedi~Dizaji et~al., 2017]{Dizaji2017}
Ghasedi~Dizaji, K., Herandi, A., Deng, C., Cai, W., and Huang, H. (2017).
\newblock Deep clustering via joint convolutional autoencoder embedding and
  relative entropy minimization.
\newblock In {\em The IEEE International Conference on Computer Vision (ICCV)}.

\bibitem[Hadsell et~al., 2006]{Hadsell:2006}
Hadsell, R., Chopra, S., and LeCun, Y. (2006).
\newblock Dimensionality reduction by learning an invariant mapping.
\newblock In {\em Proceedings of the 2006 IEEE Computer Society Conference on
  Computer Vision and Pattern Recognition - Volume 2}, CVPR '06, pages
  1735--1742.

\bibitem[Hu et~al., 2017]{Weihua2017}
Hu, W., Miyato, T., Tokui, S., Matsumoto, E., and Sugiyama, M. (2017).
\newblock Learning discrete representations via information maximizing
  self-augmented training.
\newblock In Precup, D. and Teh, Y.~W., editors, {\em Proceedings of the 34th
  International Conference on Machine Learning}, volume~70, pages 1558--1567.

\bibitem[Ji et~al., 2019]{xuIIC2018}
Ji, X., Henriques, J.~F., and Vedaldi, A. (2019).
\newblock Invariant information distillation for unsupervised image
  segmentation and clustering.
\newblock {\em arXiv preprint arXiv: 1807.06653}.

\bibitem[Jiang et~al., 2017]{zheng2016variational}
Jiang, Z., Zheng, Y., Tan, H., Tang, B., and Zhou, H. (2017).
\newblock Variational deep embedding: An unsupervised and generative approach
  to clustering.
\newblock In {\em Proceedings of the 26th International Joint Conference on
  Artificial Intelligence}, IJCAI'17, pages 1965--1972.

\bibitem[Leskovec, 2009]{LeskovecLDM09}
Leskovec, J. (2009).
\newblock Community structure in large networks: Natural cluster sizes and the
  absence of large well-defined clusters.
\newblock {\em Internet Mathematics}, 6(1):29--123.

\bibitem[Liu et~al., 2017]{liu_2017}
Liu, H., Han, J., Nie, F., and Li, X. (2017).
\newblock Balanced clustering with least square regression.
\newblock In {\em AAAI}, pages 2231--2237.

\bibitem[Mendoza~Quispe et~al., 2016]{MendozaQuispe:2016}
Mendoza~Quispe, A., Petitjean, C., and Heutte, L. (2016).
\newblock Extreme learning machine for out-of-sample extension in laplacian
  eigenmaps.
\newblock {\em Pattern Recognition}, 74(C):68--73.

\bibitem[Munkres, 1957]{Munkres1957Assignment}
Munkres, J.~R. (1957).
\newblock {Algorithms for the Assignment and Transportation Problems}.
\newblock {\em Journal of the Society for Industrial and Applied Mathematics},
  5(1):32--38.

\bibitem[Ng et~al., 2002]{ng2002spectral}
Ng, A.~Y., Jordan, M.~I., and Weiss, Y. (2002).
\newblock On spectral clustering: Analysis and an algorithm.
\newblock In {\em Advances in neural information processing systems}, pages
  849--856.

\bibitem[Shaham and Lederman, 2018]{Shaham2018LearningBC}
Shaham, U. and Lederman, R.~R. (2018).
\newblock Learning by coincidence: Siamese networks and common variable
  learning.
\newblock {\em Pattern Recognition}, 74:52--63.

\bibitem[Shaham et~al., 2018]{shaham2018spectralnet}
Shaham, U., Stanton, K., Li, H., Basri, R., Nadler, B., and Kluger, Y. (2018).
\newblock Spectralnet: Spectral clustering using deep neural networks.
\newblock In {\em International Conference on Learning Representations}.

\bibitem[Shi and Malik, 2000]{Shi_2000}
Shi, J. and Malik, J. (2000).
\newblock Normalized cuts and image segmentation.
\newblock {\em IEEE Trans. Pattern Anal. Mach. Intell.}, 22(8):888--905.

\bibitem[Von~Luxburg, 2007]{von2007tutorial}
Von~Luxburg, U. (2007).
\newblock A tutorial on spectral clustering.
\newblock {\em Statistics and computing}, 17(4):395--416.

\bibitem[Xie et~al., 2016]{xie2016unsupervised}
Xie, J., Girshick, R., and Farhadi, A. (2016).
\newblock Unsupervised deep embedding for clustering analysis.
\newblock In {\em Proceedings of the 33rd International Conference on
  International Conference on Machine Learning - Volume 48}, ICML'16, pages
  478--487.

\bibitem[Yang et~al., 2017]{yang2017}
Yang, B., Fu, X., Sidiropoulos, N.~D., and Hong, M. (2017).
\newblock Towards k-means-friendly spaces: Simultaneous deep learning and
  clustering.
\newblock In {\em Proceedings of the 34th International Conference on Machine
  Learning-Volume 70}, pages 3861--3870. JMLR. org.

\bibitem[Zhang and Rohe, 2018]{ZhangR18}
Zhang, Y. and Rohe, K. (2018).
\newblock Understanding regularized spectral clustering via graph conductance.
\newblock In {\em NeurIPS}, pages 10654--10663.

\end{thebibliography}
\bibliographystyle{apalike}

\medskip

\end{document}